\documentclass[review]{elsarticle}

\usepackage{lineno,hyperref}
\usepackage{amsmath}
\usepackage{times}
\usepackage{epsfig}
\usepackage{graphicx}
\usepackage{amsmath}
\usepackage{amssymb}
\usepackage{subfigure}
\usepackage{caption}
\usepackage{algorithm}
\usepackage{algorithmic}
\usepackage{float}
\usepackage{amsmath}
\usepackage{mathrsfs}

\journal{Journal of \LaTeX\ Templates}

\begin{document}

\begin{frontmatter}

\title{High-order structure preserving graph neural network for few-shot learning}

\author[mymainaddress]{Guangfeng Lin\corref{mycorrespondingauthor}}
\cortext[mycorrespondingauthor]{Corresponding author}
\ead{lgf78103@xaut.edu.cn}


\author[mymainaddress]{Ying Yang}
\author[mymainaddress1]{Yindi Fan}
\author[mymainaddress]{Xiaobing Kang}
\author[mymainaddress]{Kaiyang Liao}
\author[mymainaddress]{Fan Zhao}

\address[mymainaddress]{Information Science Department, Xi'an University of Technology,\\
 5 South Jinhua Road, Xi'an, Shaanxi Province 710048, PR China}
\address[mymainaddress1]{Software Technology Department,Shaanxi College of Communication Technology,\\
 19 Wenjing Road, Xi'an, Shaanxi Province 710018, PR China}

\begin{abstract}
Few-shot learning can find the latent structure information between the prior knowledge and the queried data by the similarity metric of meta-learning to construct the discriminative model for recognizing the new categories with the rare labeled samples. Most existing methods try to model the similarity relationship of the samples in the intra tasks, and generalize the model to identify the new categories. However, the relationship of samples between the separated tasks is difficultly considered because of the different metric criterion in the respective tasks. In contrast, the proposed high-order structure preserving graph neural network(HOSP-GNN) can further explore the rich structure of the samples to predict the label of the queried data on graph that enables the structure evolution to explicitly discriminate the categories by iteratively updating the high-order structure relationship (the relative metric in multi-samples,instead of pairwise sample metric) with the manifold structure constraints. HOSP-GNN can not only mine the high-order structure for complementing the relevance between samples that may be divided into the different task in meta-learning, and but also generate the rule of the structure updating by manifold constraint. Furthermore, HOSP-GNN doesn't need retrain the learning model for recognizing the new classes, and HOSP-GNN has the well-generalizable high-order structure for model adaptability. Experiments show that HOSP-GNN outperforms the state-of-the-art methods on supervised and semi-supervised few-shot learning in three benchmark datasets that are miniImageNet, tieredImageNet and FC100.
\end{abstract}


\end{frontmatter}


\section{Introduction}
Visual content recognition and understanding have greatly made progress based on the advances of deep learning methods that construct the discriminative model by training large-scale labeled data. In fact, two reasons limit the current deep learning methods for efficiently learning new categories. One is that human annotation cost is high for large-scale data (for example, thousands of the diversity samples in the same category and hundreds of the various categories in one cognition domain), the other is that the rare samples of some categories are not enough for the discriminative model training. Therefore, it is still a challenge question that the discriminative model is learned from the rare samples of the categories. To solve this question, few-shot learning \cite{vinyals2016matching} \cite{snell2017prototypical} \cite{finn2017model} \cite{RaviL17} \cite{sung2018learning} \cite{qi2018low} \cite{SatorrasE18} \cite{kim2019edge} \cite{lee2019meta} \cite{sun2019meta} \cite{peng2019few} \cite{chen2019a}proposed from the inspiration of human visual system has been an attracted research to generalize the learning model to new classes with the rare samples of each novel category by feature learning \cite{ghiasi2018dropblock} \cite{wu2018unsupervised} \cite{donahue2019large} \cite{8955914} \cite{8948295} \cite{saikia2020optimized}or meta-learning \cite{chen2019a} \cite{ravichandran2019few} \cite{dhillon2019baseline}\cite{fei2020meta} \cite{zhang2020rethinking} \cite{luo2019learning}. Feature learning emphasises on feature generation and extraction model construction based on invariance transfer information, while meta-learning focuses on the relevance model between the samples for mining the common relationship of data samples by the episode training.

\begin{figure*}[ht]
  \begin{center}
\includegraphics[width=1\linewidth]{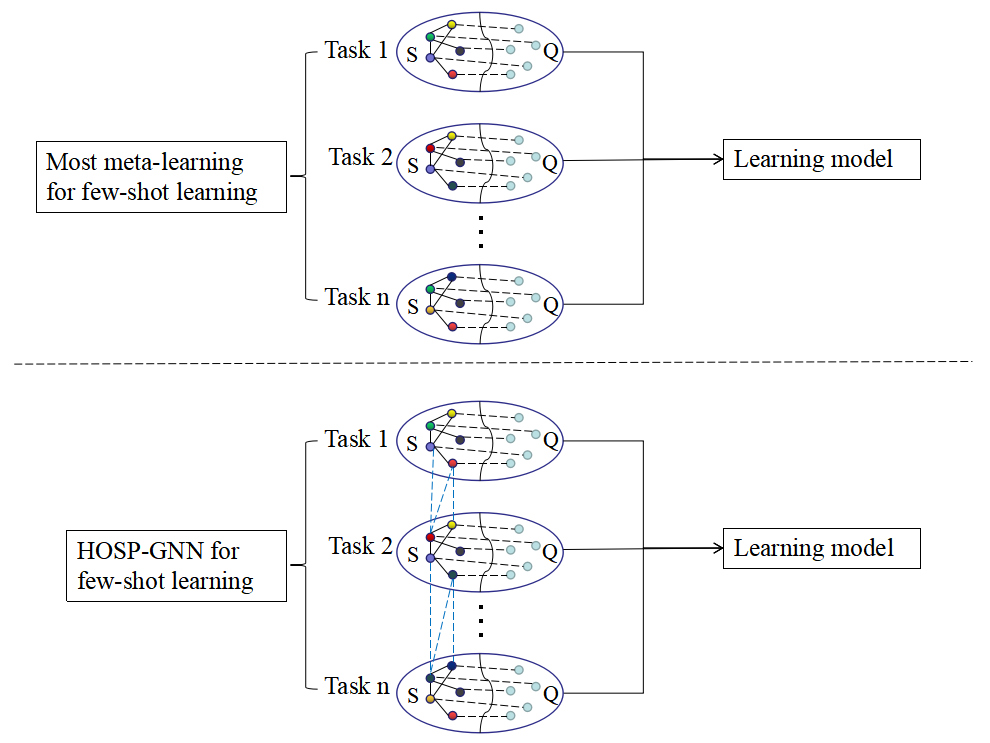}
\end{center}
\vspace{-0.2in}
 \caption{The illustration of the difference between HOSP-GNN and the most meta-learning for few-shot learning.$S$ stands for support set; $Q$ is query set;the different color circles describe the labeled samples of the different classes in $S$; the gray color circles represent unlabeled samples in $Q$;the black solid lines between circles show the structure relationship of the labeled samples;the black dot lines between circles are the predicted structure relationship between labeled and unlabeled samples; the blue dot lines between circles across tasks indicate the latent high-order structure of samples.}
  \label{fig-1}
 \end{figure*}

Meta-learning can transfer the available knowledge between the collection of the separated tasks, and propagate the latent structure information to enhance the model generalization and to avoid the model overfitting. Therefore, meta-learning is one of most promising directions for few-shot learning. However, meta-learning is constructed based on the large-scale separated tasks, and each task have the respective metric criterion that causes the gap of the transfer information between the samples of the separated tasks(the details in figure \ref{fig-2}). Although existing methods can relieve this gap to a certain extend by the same sample filling into the different tasks, it is still difficult to build the approximated metric criterion of the different tasks for efficiently information transfer and propagation. Therefore, we present HOSP-GNN that attempts to construct the approximated metric criterion by mining high-order structure and updates these metric values between samples by constraining data manifold structure for few-shot learning. Figure \ref{fig-1} illustrates the difference between HOSP-GNN and the most meta-learning for few-shot learning conceptually.

Our contributions mainly have two points as follow.
\begin{itemize}
\item One is to find the high-order structure for bridging the gap between the metric criterion of the separated tasks. The importance of this point is to balance the consistence of the same samples in the different tasks, and to enhance the transferability of the similar structure in the learning model.
\item Another is to smooth the structure evolution for improving the propagation stability of the model transfer by manifold structure constraints. This point try to minimize the difference of the transformation projection between the similar samples, and to maximize the divergence of the transformation projection between the dissimilar samples for efficiently preserving the graph structure learning of data samples.
\end{itemize}

\section{Related Works}
In recent few-shot learning, there mainly are two kinds of methods according to the different learning focuses. One is feature learning based on the data representation of model extraction, and another is meta-learning based on the metric relationship of model description.

\subsection{Feature Learning}
Feature learning \cite{cubuk2019autoaugment} \cite{cubuk2019randaugment} \cite{lim2019fast} \cite{ghiasi2018dropblock} \cite{wu2018unsupervised} \cite{donahue2019large}for few-shot learning expects to inherit and generalize the well characteristics of the pre-train model based on large-scale samples training for recognizing new classes with few samples.

Because few samples often can not satisfy the necessary of the whole model training, the recent representative methods usually optimize the part parameters or structure of the pre-trained model by few samples for feature learning. For example,Bayesian optimization to Hyperband (BOHB) optimizes hyper-parameters by searching the smaller parameter space to maximize the validation performance for generic feature learning \cite{saikia2020optimized}; Geometric constraints fine-tune the parameters of one network layer with a few training samples for extracting the discriminative features for the new categories \cite{8948295}; Bidirectional projection learning (BPL) \cite{8955914} utilizes semantic embedding to synthesize the unseen classes features for obtaining enough samples features by competitive learning. These methods attempt to find the features invariance by partly fine-tuning the pre-trained model with the different constraints for recognizing the new classes with the few instances.

However, these methods can not explicitly formulate the metric rules for learning the discriminative model between new categories, moreover, these methods need retrain the model to adapt the distribution of new categories. It can lead to the degraded classification performance for few-shot learning and the more complicated optimization strategy in validation and test phases.

\subsection{Meta-Learning}
\label{ML}
Meta-learning \cite{vinyals2016matching} \cite{sung2018learning} \cite{chen2019a}\cite{gidaris2018dynamic} \cite{oreshkin2018tadam} \cite{ravichandran2019few} \cite{dhillon2019baseline}for few-shot learning tries to construct the relevances between samples in the base classes for generalizing the model to the new classes. These methods can learn the common structure relationship between samples by training on the collection of the separated tasks.In terms of the coupling between the model and the data, meta-learning can mainly be divided into two groups.

One group is the model optimization to quickly fit the distribution of new categories. Typical methods attempt to update the model parameters or optimizer for this purpose. For instance, meta-learner long short-term memory (LSTM) \cite{RaviL17} can update the model parameters to initialize the classifier network for the quick training convergence in the few samples of each classes; Model-agnostic meta-learning (MAML) \cite{finn2017model} can train the small gradient updating based on few learning data from a new task to obtain the well generalization performance; Latent embedding optimization(LEO)\cite{rusu2019metalearning}
can learn the latent generative representation of model parameters based on data dependence to decouple the gradient adaptation from the high-dimension parameters space.

Another group is metric learning to describe the structure relationship of the samples between support and query data for directly simulating the similarity metric of the new categories. The recent methods trend to enhance the metric structure by constraint information for the better model generalization in the new categories. For example,edge-labeling graph neural network (EGNN)\cite{kim2019edge} can update graph structure relationship to directly exploit the intra-cluster similarity and the inter-cluster dissimilarity by iterative computation; Meta-learning across meta-tasks (MLMT) \cite{fei2020meta} can explore their relationships between the random tasks by meta-domain adaptation or meta-knowledge distillation for boosting the performance of existing few-shot learning methods; Absolute-relative Learning (ArL)\cite{zhang2020rethinking} can both consider the class concepts and the similarity learning to complement their structure relationship for improving the recognition performance of the new categories; Continual meta-learning approach with Bayesian graph neural networks(CML-BGNN) \cite{luo2019learning} can implement the continual learning of a sequence of tasks to preserve the intra-task and inter-task correlations by message-passing and history transition.

In recent work, meta-learning based on metric learning shows the promising performance for recognizing the new categories with the few samples. These methods initially focus on the structure relation exploitation between support set and query set by modeling metric distances, and subsequent works further mine the relevance by mimicking the dependence between the separated tasks for enhancing the discrimination of the new categories. However, these methods depend on the projection loss between the seen and unseen classes \cite{fei2020meta} or Bayesian inference based on low-order structure (the metric of the pairwise data) \cite{luo2019learning} for considering the structure relationship between the intra or inter tasks. It is difficult to describe the latent high-order structure from the global observation. Therefore, the proposed HOSP-GNN expects to capture the high-order structure relationship based on samples metric for naturally correlating the relevance between the intra or inter tasks for improving the performance of few-shot learning.

\section{High-order structure preserving graph neural network}
Few-shot classification attempts to learn a classifier model for identifying the new classes with the rare samples. $C_{e}$ or $C_{n}$ respectively stands for a existing classes set with the large samples or a new classes set with the rare samples, and $C_{e}\bigcap C_{n}=\emptyset$,but they belong to the same cognise domain. The existing classes data set $D_{e}=\{(x_{i},y_{i})|y_{i} \in C_{e},  i=1,...,|D_{e} |\}$, where $x_{i}$ indicates the $i$-th image with the class label $y_{i}$, $|D_{e}|$ is the number of the elements in $D_{e}$. Similarly, the new classes data set $D_{n}=\{(x_{i},y_{i})|y_{i} \in C_{n},  i=1,...,|D_{n} |\}$, where $x_{i}$ indicates the $i$-th image with the class label $y_{i}$, $|D_{n}|$ is the number of the elements in $D_{n}$. If each new class includes $K$ labeled samples, the new classes data set is $K$-shot sample set. In other word, $|D_{n}|=K|C_{n}|$, where $|C_{n}|$ is the number of the elements in $C_{n}$. Few-shot learning is to learn the discriminative model from $D_{n}$ to predict the label of the image sample in the test set $D_{t}$ that comes from $C_{n}$ and $D_{n} \bigcap D_{t}=\emptyset$.
\subsection{Meta-learning for few-shot learning based on graph neural network}
In meta-learning, the classifier model can be constructed based on the collection of the separated tasks $\mathcal{T}=\{S,Q\}$ that contains a support set $S$ from the labeled samples in $D_{n}$ and a query set $Q$ from unlabeled samples in $D_{t}$. To build the learning model for few-shot learning, $S$ includes $K$ labeled samples and $N$ classes, so this situation is called $N$-way-$K$-shot few-shot classification that is to distinguish the unlabeled samples from $N$ classes in $Q$.

In practise, few-shot classification often faces the insufficient model learning based on the new classes data set $D_{n}$ with the rare labeled samples and $D_{t}$ with unlabeled samples. In this situation, the model difficultly identifies the new categories. Therefore, many methods usually draw support from the transfer information of $D_{e}$ with a large labels samples to enhance the model learning for recognizing the new classes. Episodic training \cite{sung2018learning} \cite{kim2019edge} is an efficient meta-learning for few-shot classification. This method can mimic $N$-way-$K$-shot few-shot classification in $D_{n}$ and $D_{t}$ by randomly sampling the differently separated tasks in $D_{e}$ as the various episodics of the model training. In each episode, $\mathcal{T}_{ep}=(S_{ep},Q_{ep})$ indicates  the separated tasks with $N$-way-$K$-shot $T$ query samples, where the support set $S_{ep}=\{(x_{i},y_{i})|y_{i}\in C_{ep}, i=1,...,N\times K\}$, the query set  $Q_{ep}=\{(x_{i},y_{i})|y_{i}\in C_{ep}, i=1,...,N\times T\}$, $S_{ep}\cap Q_{ep}=\emptyset$, and the class number $|C_{ep}|=N$. In the training phase, the class set $C_{ep}\in C_{e}$, while in test phase, the class set $C_{ep}\in C_{n}$. Many episodic tasks can be randomly sampled from $D_{e}$ to simulate $N$-way-$K$-shot learning for training the few-shot model, whereas the learned model can test the random tasks from $D_{n}$ for few-shot classification by $N$-way-$K$-shot fashion. If we construct a graph $G_{ep}=(\mathcal{V}_{ep},\mathcal{E}_{ep},\mathcal{T}_{ep})$ (here, $\mathcal{V}_{ep}$ is the vertex set of the image features in $\mathcal{T}_{ep}$, and $\mathcal{E}_{ep}$ is the edge set between the image features in $\mathcal{T}_{ep}$.) for describing the sample structure relationship in each episodic task, meta-learning for few-shot learning based on $L$ layers graph neural network can be reformulated by the cross-entropy loss $L_{ep}$ as following.
\begin{align}
\label{Eq1}
 \begin{aligned}
  L_{ep}&=-\sum_{l=1}^{L}\sum_{(x_{i},y_{i})\in Q_{ep}}y_{i}\log (h_{W}^{l}(f(x_{i},W_{f});S_{ep},G_{ep}))\\
  &=-\sum_{l=1}^{L}\sum_{(x_{i},y_{i})\in Q_{ep}}y_{i}\log (\hat{y_{i}^{l}})
 \end{aligned}
\end{align}

\begin{align}
\label{Eq1-1}
 \begin{aligned}
  \hat{y_{i}^{l}}=softmax(\sum_{j\neq i~~and~~c\in C_{ep}}e_{ij}^{l}\delta(y_{i}=c))
 \end{aligned}
\end{align}

here, $\hat{y_{i}^{l}}$ is the estimation value of $y_{i}$ in $l$th layer;$e_{ij}^{l}$ is edge feature of the $l$th layer in graph $G_{ep}$; $\delta(y_{i}=c)$ is equal one when $y_{i}=c$ and zero otherwise;$f(\bullet)$ with the parameter set $W_{f}$ denotes the feature extracting function or network shown in Figure \ref{fig-4}(a); $h_{W}^{l}(f(x_{i});S_{ep},G_{ep})$ indicates few-shot learning model in the $l$th layer by training on $S_{ep}$ and $G_{ep}$, and $W$ is the parameter set of this model. This few-shot learning model can be exploited by the meta-training minimizing the loss function \ref{Eq1}, and then recognize the new categories with the rare samples.

\subsection{High-order structure description}
\begin{figure*}[ht]
  \begin{center}
\includegraphics[width=1\linewidth]{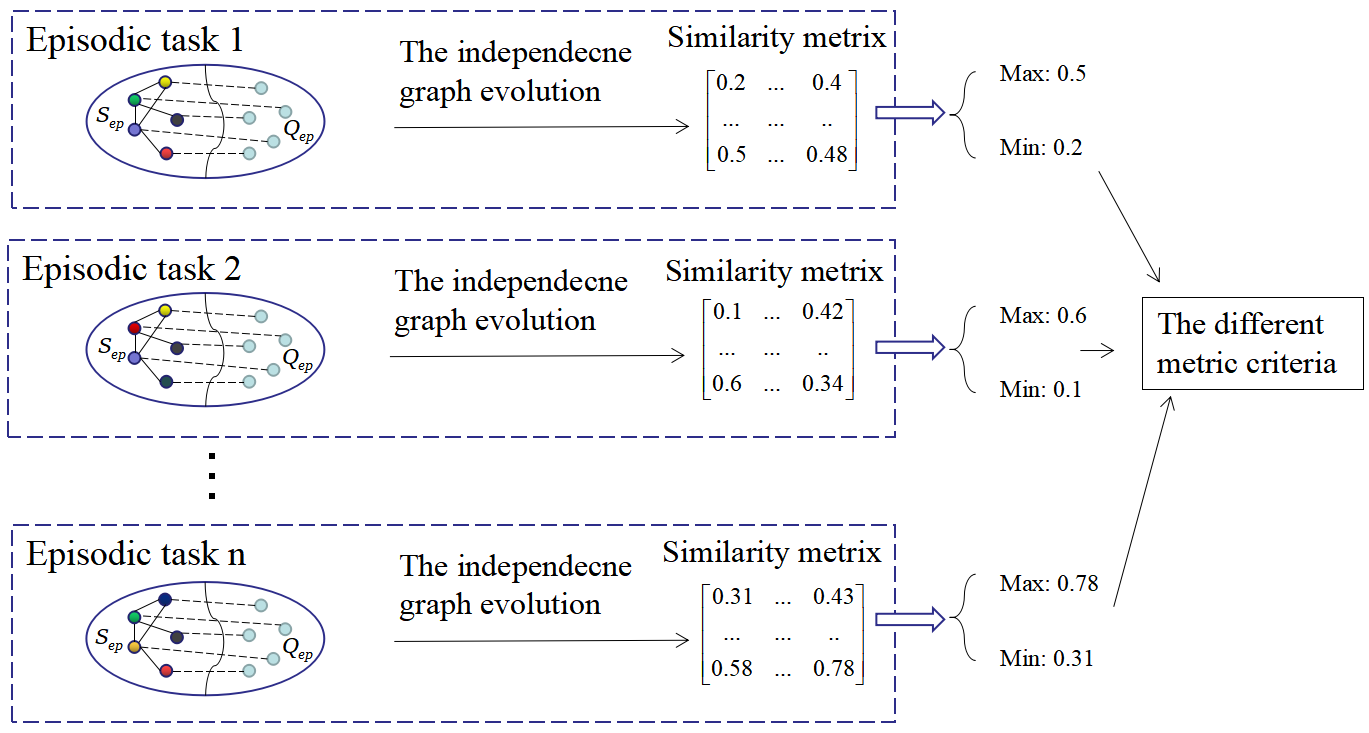}
\end{center}
\vspace{-0.2in}
 \caption{The difference between the metric criteria of the episodic tasks. In each episodic training and testing, $S_{ep}$ stands for support set; $Q_{ep}$ is query set;the different color circles describe the labeled sample of the different classes in $S_{ep}$; the gray color circles represent unlabeled samples in $Q_{ep}$;the black solid lines between circles show the structure relationship of the labeled samples;the black dot lines between circles are the predicted structure relationship between labeled and unlabeled samples.}
  \label{fig-2}
 \end{figure*}

In few-shot learning based on graph neural network, the evolution and generation of the graph plays a very important role for identifying the different classes. In each episodic task of meta-learning, existing methods usually measure the structure relationship of the samples by pairwise way, and an independence metric space with the unique metric criteria is formed by the similarity matrix in graph. In many episodic tasks training, the various  metric criteria lead to the divergence between the different samples structure relationship in Figure \ref{fig-2}. It is the main reason that the unsatisfactory classification of the new categories.

To reduce the difference between the metric criteria of the episodic tasks, we attempt to explore the high-order structure of the samples by building the latent connection. The traditional pairwise metric loses the uniform bench marking because of the normalization of the sample separation in independence tasks. However, the absolutely uniform bench marking is difficult to build the high-order structure relation between the samples of the different tasks. Therefore, \textbf{we define the relative metric graph of multi-samples in a task as high-order structure relation}, and the same samples by random falling into the independence task make this relative metric relationship widely propagate to the other samples for approximating to the uniform bench marking under the consideration with the interaction relationship between the episodic tasks.

\begin{figure*}[ht]
  \begin{center}
\includegraphics[width=1\linewidth]{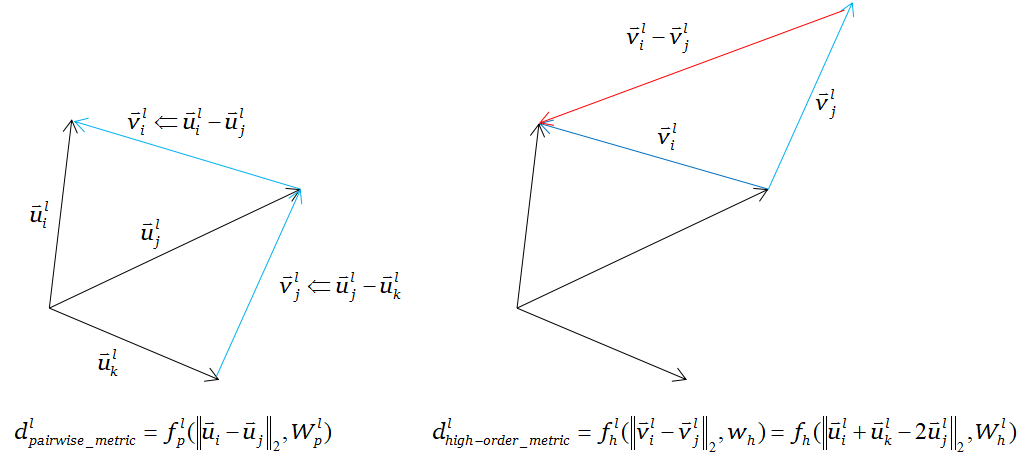}
\end{center}
\vspace{-0.2in}
 \caption{The relationship between pairwise metric $d_{pairwise\_metric}^{l}$(left figure) and high-order metric $d_{high-order\_metric}^{l}$(right figure) in $l$th layer. $f_{p}^{l}(\bullet)$ and $W_{p}^{l}$ respectively are pairwise metric network projection and parameter set in $l$th layer, while $f_{h}^{l}(\bullet)$ and $W_{h}^{l}$ respectively are high-order metric network projection and parameter set in $l$th layer.The black vector indicates the original vertex, the blue vector is the low-order metric vector ( the relative metric based on the original vertex), and the red vector stands for the high-order metric vector.}
  \label{fig-3}
 \end{figure*}

More concretely, the relative metric graph $\hat{G}_{ep}=(\hat{\mathcal{V}}_{ep},\hat{\mathcal{E}}_{ep},\mathcal{T}_{ep})$, where $\mathcal{T}_{ep}=\{(x_{i},y_{i})|(x_{i},y_{i})\in S_{ep}~~or~~(x_{i},y_{i})\in Q_{ep}, y_{i}\in C_{ep},S_{ep}\bigcap Q_{ep}=\emptyset, i=1,...,N\times (K+T)\}$, the vertex set $\hat{\mathcal{V}}_{ep}=\{v_{i}|i=1,...,N\times (K+T)\}$, the edge set $\hat{\mathcal{E}}_{ep}=\{e_{ij}|i=1,...,N\times (K+T)~~and~~j=1,...,N\times (K+T)\}$. To describe the relative relationship between features, we can build $L$ layers graph neural network for learning edge feature $e_{ij}^{l}$  (graph structure relationship) and feature representation $v_{i}^{l}$ in each layer, where $l=0,...,L$. In the initial layer, each vertex feature $v_{i}^{0}$ can be computed by feature difference as following.
\begin{align}
\label{Eq2-1}
 \begin{aligned}
&u_{i}^{0}=f(x_{i}),~~~~~~i=1,...,N\times (K+T),
 \end{aligned}
\end{align}

\begin{align}
\label{Eq2}
 \begin{aligned}
v_{i}^{0}=\left\{
  \begin{aligned}
&u_{i}^{0}-u_{i+1}^{0},~~~~~~i=1,...,N\times (K+T)-1, \\
&u_{i}^{0}-u_{1}^{0},~~~~~~~~~~i=N\times (K+T),
\end{aligned}
\right.
 \end{aligned}
\end{align}
here, $f(\bullet)$ is the feature extracting network shown in Figure \ref{fig-4}(a). The vertex can represented by two ways. One is that the initial vertex feature $u_{i}^{0}$ is described by the original feature. Another is that $v_{i}^{0}$  is a relative metric based on $u_{i}^{0}$ in $0$th layer. We expect to construct the higher order structure $e_{ij1}^{l}$ (the first dimension value of edge feature between vertex $i$ and $j$ in $l$ layer) based on this relative metric for representing edge feature under the condition with the pairwise similarity structure $e_{ij2}^{l}$ and dissimilarity structure $e_{ij3}^{l}$(these initial value of $0$ layer is defined by the labeled information of $S_{ep}$ in Equation \ref{Eq3}). Therefore, the initial edge feature can be represented by the different metric method as following.

\begin{align}
\label{Eq3}
 \begin{aligned}
e_{ij}^{0}=\left\{
  \begin{aligned}
& [e_{ij1}^{0}~||~ e_{ij2}^{0}=1~||~ e_{ij3}^{0}=0],~~~~y_{i}=y_{j}~~and~~(x_{i},y_{i})\in S_{ep}, \\
& [e_{ij1}^{0}~||~ e_{ij2}^{0}=0~||~ e_{ij3}^{0}=1],~~~~y_{i}\neq y_{j}~~and~~(x_{i},y_{i})\in S_{ep}, \\
& [e_{ij1}^{0}~||~ e_{ij2}^{0}=0.5~||~ e_{ij3}^{0}=0.5],~~~~ otherwise,
\end{aligned}
\right.
 \end{aligned}
\end{align}
here, $||$ is concatenation symbol, $e_{ij1}^{0}$ can be calculated by the metric distance of the  difference in Equation \ref{Eq4}, and $e_{ij1}^{l}$ can be updated by Equation \ref{Eq7}. It shows the further relevance between the relative metric, and indicates the high-order structure relation of the original features.

\begin{align}
\label{Eq4}
 \begin{aligned}
e_{ij1}^{0}=1-\parallel v_{i}^{0}-v_{j}^{0} \parallel_{2}/\sum_{k}\parallel v_{i}^{0}-v_{k}^{0} \parallel_{2},~~~~(x_{i},y_{i})\in S_{ep}\bigcup Q_{ep},
 \end{aligned}
\end{align}
Figure \ref{fig-3} shows the relationship between pairwise metric and high-order metric in $l$th layer, and  the high-order metric involves any triple vertex features $u_{i}^{l}$,$u_{j}^{l}$ and $u_{k}^{l}$ in $\hat{G}_{ep}$ in each task. In these features, $u_{j}^{l}$ is a benchmark feature that is randomly sampled by the separated tasks. The common benchmark feature can reduce the metric difference between samples of the separated tasks.

\subsection{High-order structure preserving}
\begin{figure*}[hbp]
  \begin{center}
\includegraphics[width=1\linewidth]{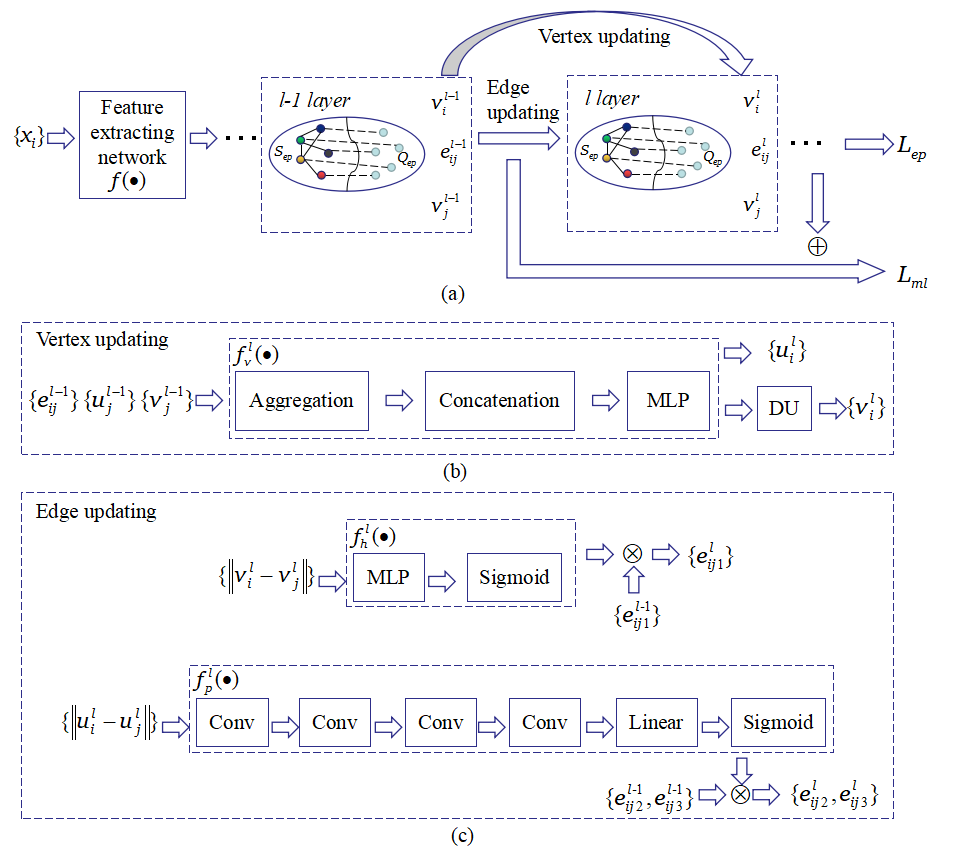}
\end{center}
\vspace{-0.2in}
 \caption{The network architecture of the proposed HOSP-GNN.(a) is the total network structure, (b) and (c) respectively are vertex and edge updating network in (a). MLP is a multilayer perceptron; DU indicates the difference unit for the relative metric; Conv stands for a convolutional block that includes $96$ channels of $1\times 1$ convolution kernel, batch normalization unit, and LeakReLU unit;$S_{ep}$ stands for support set; $Q_{ep}$ is query set;the different color circles describe the labeled samples of the different classes in $S_{ep}$; the gray color circles represent unlabeled samples in $Q_{ep}$;the black solid lines between circles show the structure relationship of the labeled samples;the black dot lines between circles are the predicted structure relationship between labeled and unlabeled samples;$L_{ep}$ is the loss metric between the real labels and the predicted labels; $L_{ml}$ is the loss metric between high structures layer by layer;$v_{i}^{l}$ is the $i$th vertex feature in the $l$th layer of graph;$e_{ij}^{l}$ is the edge feature between the vertex $i$ and $j$ in the $l$th layer of graph;$f(\bullet)$ denotes the feature extracting network;$f_{v}^{l}(\bullet)$ indicates the vertex feature updating network in the $l$th layer; $f_{h}^{l}(\bullet)$ denotes the high-order metric network between vertex features in the $l$th layer; $f_{p}^{l}(\bullet)$ stands for the pairwise metric network between vertex features in the $l$th layer.}
  \label{fig-4}
 \end{figure*}

HOSP-GNN can construct $L$ layers graph neural network for evolving the graph structure by updating the vertex and edge features. Moreover, we expect to preserve the high-order structure layer by layer for learning the discriminative structure between samples in the separated tasks. $l=1,...,L$ is defined as the layer number. In detail, $u_{i}^{l}$ can be updated by $u_{i}^{l-1}$,$v_{i}^{l-1}$ and $e_{ij}^{l-1}$ in Equation \ref{Eq5}, while $e_{ij}^{l}$ can be updated by$u_{i}^{l-1}$, $v_{i}^{l-1}$ and $e_{ij}^{l-1}$ in Equation \ref{Eq7},\ref{Eq8} and \ref{Eq9}.

\begin{align}
\label{Eq5}
 \begin{aligned}
u_{i}^{l}=f_{v}^{l}([\sum_{j}\tilde{e}_{ij1}^{l-1}v_{j}^{l-1}~||~\sum_{j}\tilde{e}_{ij2}^{l-1}u_{j}^{l-1}~||~\sum_{j}\tilde{e}_{ij3}^{l-1}u_{j}^{l-1}], W_{v}^{l}),
 \end{aligned}
\end{align}

\begin{align}
\label{Eq5-1}
 \begin{aligned}
v_{i}^{l}=\left\{
  \begin{aligned}
&u_{i}^{l}-u_{i+1}^{l},~~~~~~i=1,...,N\times (K+T)-1, \\
&u_{i}^{l}-u_{1}^{l},~~~~~~~~~~i=N\times (K+T),
\end{aligned}
\right.
 \end{aligned}
\end{align}
here,$||$ is concatenation symbol, $\tilde{e}_{ijk}^{l-1}=e_{ijk}^{l-1}/\sum_{k}e_{ijk}^{l-1}$ ($k=1,2,3$), and $f_{v}^{l}(\bullet)$ is the vertex feature updating network shown in Figure \ref{fig-4}(b),and $W_{v}^{l}$ is the network parameters in $l$th layer. This updating process shows that the current vertex feature is the aggregative transformation of the previous layer vertex and edge feature in the different metrics, and can propagate the representation information under the consideration with edge feature (high-order structure information) layer by layer evolution. In \ref{Eq5}, high-order structure influences the vertex representation by transforming aggregation computation, but can not efficiently transfer layer by layer. Therefore, we expect to preserve high-order structure layer by layer by updating edge features. According to manifold learning \cite{he2004locality} and structure fusion\cite{Lin2014146}, structure information (the similarity relationship of samples) can be held from the original space to the projection space by minimizing the metric difference of these spaces. Similarly, high-order evolution based on graph neural network may obey the same rule for computing edge feature of each layer with the vertex feature updating. Therefore, we can construct the manifold loss by layer-by-layer computation for constraining the model optimization.

\begin{align}
\label{Eq6}
 \begin{aligned}
L_{ml}=&\sum_{i,j,l}f_{h}^{l}(\|v_{i}^{l}-v_{j}^{l}\|_{2},W_{h}^{l})e_{ij1}^{l-1}+\\
&\sum_{i,j,l}f_{p}^{l}(\|u_{i}^{l}-u_{j}^{l}\|_{2},W_{p}^{l})e_{ij2}^{l-1}+\\
&\sum_{i,j,l}(1-f_{p}^{l}(\|u_{i}^{l}-u_{j}^{l}\|_{2},W_{h}^{l}))e_{ij3}^{l-1},
\end{aligned}
\end{align}
here,$L_{ml}$ is the loss of the manifold structure in the different layer and metric method (The first term is the manifold constrain for high-order structure, while the second and third terms are respectively the manifold constrain for similarity and dissimilarity); $f_{h}^{l}(\bullet)$ is the high-order metric network in Figure \ref{fig-4}(c) between vertex features, and $W_{h}^{l}$ is the parameter set of this network in $l$th layer; $f_{p}^{l}(\bullet)$ is the pairwise metric network in Figure \ref{fig-4}(c) between vertex features, and $W_{p}^{l}$ is it's parameter set in $l$th layer. \ref{Eq6} shows that the different manifold structures between layers can be preserved for minimizing $L_{ml}$. The edge updating based on high-order structure preserving is as following.

\begin{align}
\label{Eq7}
 \begin{aligned}
\bar{e}_{ij1}^{l}=\frac{f_{h}^{l}(\|v_{i}^{l}-v_{j}^{l}\|_{2},W_{h}^{l})e_{ij1}^{l-1}}{\sum_{k}f_{h}^{l}(\|v_{i}^{l}-v_{k}^{l}\|_{2},W_{h}^{l})e_{ik1}^{l-1}/\sum_{k}e_{ik1}^{l-1}},
\end{aligned}
\end{align}

\begin{align}
\label{Eq8}
 \begin{aligned}
\bar{e}_{ij2}^{l}=\frac{f_{p}^{l}(\|u_{i}^{l}-u_{j}^{l}\|_{2},W_{p}^{l})e_{ij2}^{l-1}}{\sum_{k}f_{p}^{l}(\|u_{i}^{l}-u_{k}^{l}\|_{2},W_{p}^{l})e_{ik2}^{l-1}/\sum_{k}e_{ik2}^{l-1}},
\end{aligned}
\end{align}

\begin{align}
\label{Eq9}
 \begin{aligned}
\bar{e}_{ij3}^{l}=\frac{(1-f_{p}^{l}(\|u_{i}^{l}-u_{j}^{l}\|_{2},W_{p}^{l}))e_{ij3}^{l-1}}{\sum_{k}(1-f_{p}^{l}(\|u_{i}^{l}-u_{k}^{l}\|_{2},W_{p}^{l}))e_{ik3}^{l-1}/\sum_{k}e_{ik3}^{l-1}},
 \end{aligned}
\end{align}

\begin{align}
\label{Eq10}
 \begin{aligned}
e_{ij}^{l}=\bar{e}_{ij}^{l}/\|\bar{e}_{ij}^{l}\|_{1}.
 \end{aligned}
\end{align}
Therefore, The total loss $L_{total}$ of the whole network includes $L_{ep}$ and $L_{ml}$.

\begin{align}
\label{Eq11}
 \begin{aligned}
 L_{total}=L_{ep}+\lambda L_{ml},
 \end{aligned}
\end{align}
here, $\lambda$ is the tradeoff parameter for balancing the influence of the different loss. Figure \ref{fig-4} shows the network architecture of the proposed HOSP-GNN.

To indicate the inference details of HOSP-GNN, algorithm \ref{algHOSP-GNN} shows the pseudo code of the proposed HOSP-GNN for predicting the labels of the rare samples. This algorithm process contains four steps. The first step (line 1 and line 2) initializes the vertex feature and the edge feature. The second step (line 4 and line 5) updates the vertex features layer by layer. The third step (from line 6 to line 8) updates the edge features layer by layer. The forth step (line 9) predicts the labels of the query samples.
\begin{algorithm}[ht]
  \caption{The inference of the HOSP-GNN for few-shot learning}
 \begin{algorithmic}[1]
 \label{algHOSP-GNN}
\renewcommand{\algorithmicrequire}{\textbf{Input:}}
\renewcommand{\algorithmicensure}{\textbf{Output:}}
\renewcommand{\algorithmicreturn}{\textbf{Iteration:}}
   \REQUIRE \textit{Graph}, $\hat{G}_{ep}=(\hat{\mathcal{V}}_{ep},\hat{\mathcal{E}}_{ep},\mathcal{T}_{ep})$, where $\mathcal{T}_{ep}=\{(x_{i},y_{i})|(x_{i},y_{i})\in S_{ep}~~or~~x_{i}\in Q_{ep}, y_{i}\in C_{ep},S_{ep}\bigcap Q_{ep}=\emptyset, i=1,...,N\times (K+T)\}$,$\hat{\mathcal{V}}_{ep}=\{v_{i}|i=1,...,N\times (K+T)\}$, $\hat{\mathcal{E}}_{ep}=\{e_{ij}|i=1,...,N\times (K+T)~~and~~j=1,...,N\times (K+T)\}$;\\ \textit{Model parameter}, $W=\{W_{f},W_{v}^{l},W_{h}^{l}|l=1,...,L\}$
   \ENSURE The query samples of the predicted labels $\{\hat{y}_{i}^{l}|i=1,...,N\times T~~~~and~~~~l=1,...,L\}$
   \STATE Computing the initial vertex feature $v_{i}^{0}$ by feature difference in Equation \ref{Eq2}
   \STATE Computing the initial edge feature $e_{ij}^{0}$ as high-order structure in Equation \ref{Eq3}
   \FOR {$1\leq l\leq L$}
       \FOR {$1\leq i\leq N\times (K+T)$}
       \STATE Updating vertex feature $v_{i}^{l}$ by Equation \ref{Eq5}
          \FOR {$1\leq j\leq N\times (K+T)$}
          \STATE Updating edge feature $e_{ij}^{l}$ by Equation \ref{Eq7},\ref{Eq8},\ref{Eq9} and \ref{Eq10}
          \ENDFOR
       \STATE Predicting the query sample labels $\hat{y}_{i}^{l}$ by Equation\ref{Eq1-1}
       \ENDFOR
   \ENDFOR

  \end{algorithmic}
\end{algorithm}

\section{Experiment}

To evaluating the proposed HOSP-GNN, we carry out four experiments. The first experiment involves the baseline methods comparison. The second experiment conducts the state-of-the-art methods comparison. The third experiment implements semi-supervised fashion for few-shot learning. The forth experiment assesses the layer effect for graph model, and the loss influence for the manifold constraint.
\subsection{Datasets}
In experiments, we use three benchmark datasets that are miniImageNet\cite{vinyals2016matching}, tieredImageNet \cite{Mengye2018}, and FC100\cite{NIPS20187352}. In miniImageNet dataset from ILSVRC-12 \cite{ILSVRC15}, RGB images include $100$ different classes, and each class has $600$ samples. We adopt the splits configuration \cite{kim2019edge} that respectively is 64,16,and 20 classes for training, validation and testing. In tieredImageNet dataset from ILSVRC-12 \cite{ILSVRC15}, there are more than $700k$ images from  $608$ classes. Moreover, $608$ classes is collected for $34$ higher-level semantic classes, each of which has $10$ to $20$ classes. We also use the splits configuration \cite{kim2019edge} that respectively is $351$,$97$, and $160$ for training, validation and testing. Each class has about $1281$ images. In FC100 dataset from CIFAR-100\cite{Krizhevsky}, there are $100$ classes images grouped into $20$ higher-level classes. Classes respectively are divided into $60$,$20$, and $20$ for training, validation and testing. Each classes have $600$ images of size $32\times 32$. Table \ref{tab1} shows the statistics information of these datasets.

\begin{table*}[!ht]
\small
\renewcommand{\arraystretch}{1.0}
\caption{Datasets statistics information in experiments. $\sharp$ denotes the number. }
\label{tab1}
\begin{center}
\newcommand{\tabincell}[2]{\begin{tabular}{@{}#1@{}}#2\end{tabular}}
\begin{tabular}{lp{1.0cm}p{1.5cm}p{1.5cm}p{1.5cm}p{1.5cm}p{0.5cm}}
\hline
\bfseries Datasets & \bfseries \tabincell{l}{$\sharp$ Classes } & \bfseries \tabincell{l}{$\sharp$ training\\ classes} & \bfseries \tabincell{l}{$\sharp$ validation \\classes} & \bfseries \tabincell{l}{$\sharp$ testing \\classes} &\bfseries \tabincell{l}{$\sharp$ images}\\
\hline \hline
miniImageNet  & $100$ &$64$& $16$ & $20$ & $60000$\\
\hline
tieredImageNet  & $608$ &$351$& $97$ & $160$ & $778848$\\
\hline
FC100  & $100$ &$60$& $20$ & $20$ & $60000$\\
\hline
\end{tabular}
\end{center}
\end{table*}

\subsection{Experimental Configuration}
Figure \ref{fig-4} describes the network architecture of the proposed HOSP-GNN in details. The feature extracting network is the same architecture in the recent works\cite{vinyals2016matching} \cite{snell2017prototypical} \cite{finn2017model} \cite{kim2019edge}, and specifically includes four convolutional blocks with $3\times 3$ kernel, one linear unit, one bach normalization and one leakReLU unit for few-shot models. Other parts of network is detailed in figure \ref{fig-4}. To conveniently compare with other methods(baseline methods and state-of-the-art methods), we set the layer number $L$ to $3$ in the proposed HOSP-GNN.

To train the proposed HOSP-GNN model, we use Adam optimizer with the learning rate $5\times 10^{-4}$ and weight decay $10^{-6}$. The mini-batch size of meta-learning task is set to $40$ or $20$ for 5-way-1-shot or 5-way-5-shot experiments. The loss coefficient $\lambda$ is set to $10^{-5}$. Experimental results in this paper can be obtained by 100K iterations training for miniImageNet and FC100, 200K iterations training for tieredImageNet.

We implement 5-way-1-shot or 5-way-5-shot experiments for evaluating the proposed method. Specifically, we averagely sample $15$ queries from each classes, and randomly generate $600$ episodes from the test set for calculating the averaged performance of the queries classes.

\subsection{Comparison with baseline approaches}
The main framework of the proposed HOSP-GNN is constructed based on edge-labeling graph neural network (EGNN)\cite{kim2019edge}. Their differences are the graph construction and the manifold constraint for model training in episodic tasks. EGNN method mainly considers the similarity and dissimilarity relationship between the pair-wise samples, but does not involve the manifold structure constraint of each layer for learning few-shot model.In contrast, HOSP-GNN tries to capture the high-order structure relationship between multi-samples ,fuses the similarity and dissimilarity relationship between the pair-wise samples, and constrains the model training by layer by layer manifold structure loss. Therefore, the base-line methods include EGNN, HOSP-GNN-H-S(the proposed HOSP-GNN only considers the high-order structure relationship and the similarity relationship),HOSP-GNN-H-D(the proposed HOSP-GNN only considers the high-order structure relationship and the dissimilarity relationship),HOSP-GNN-H (the proposed HOSP-GNN only considers the high-order structure relationship),HOSP-GNN-S (the proposed HOSP-GNN only considers the similarity relationship),and HOSP-GNN-D (the proposed HOSP-GNN only considers the dissimilarity relationship), in which H denotes the high-order structure relationship, S strands for the similarity relationship, and D represents the dissimilarity relationship.

\begin{table}[!ht]
\small
\renewcommand{\arraystretch}{1.0}
\caption{Comparison of the methods related the high-order structure (HOSP-GNN,HOSP-GNN-H-S,HOSP-GNN-H-D,and HOSP-GNN-H)with baseline methods (EGNN,HOSP-GNN-S,and HOSP-GNN-D) for  5-way-1-shot learning. Average accuracy (\%)of the query classes is reported in random episodic tasks.}
\label{tab2}
\begin{center}
\newcommand{\tabincell}[2]{\begin{tabular}{@{}#1@{}}#2\end{tabular}}
\begin{tabular}{l|p{2.0cm}p{2.0cm}p{2.0cm}p{2.0cm}}
\hline
\bfseries Method &\bfseries 5-way-1-shot &\bfseries &\bfseries  \\
\cline{2-4}
\bfseries  &\bfseries miniImageNet &\bfseries tieredImageNet &\bfseries FC100 \\
\hline \hline
EGNN \cite{kim2019edge}  & $52.46\pm0.45$ &$57.94\pm0.42$ & $35.00\pm0.39$ \\
\hline
HOSP-GNN-D & $52.44\pm0.43$ &$57.91\pm0.39$ & $35.55\pm0.40$ \\
\hline
HOSP-GNN-S & $52.86\pm0.41$ &$57.84\pm0.44$ & $35.48\pm0.42$ \\
\hline\hline
HOSP-GNN-H & $69.52\pm0.41$ &$91.71\pm 0.28$ & $76.24\pm0.41$ \\
\hline
HOSP-GNN-H-D & $78.82\pm0.45$ &$82.63\pm0.26$ & $82.27\pm0.44$ \\
\hline
HOSP-GNN-H-S & $88.15\pm0.35$ &$\textbf{95.39}\pm\textbf{0.20}$ & $\textbf{83.65}\pm\textbf{0.38}$ \\
\hline
HOSP-GNN & $\textbf{93.93}\pm\textbf{0.37}$ &$94.00\pm0.24$ & $76.79\pm0.46$ \\
\hline
\end{tabular}
\end{center}
\end{table}

\begin{table}[!ht]
\small
\renewcommand{\arraystretch}{1.0}
\caption{Comparison of Comparison of the methods related the high-order structure (HOSP-GNN,HOSP-GNN-H-S,HOSP-GNN-H-D,and HOSP-GNN-H) with baseline methods (EGNN,HOSP-GNN-S,and HOSP-GNN-D) for  5-way-5-shot learning. Average accuracy (\%)of the query classes is reported in random episodic tasks.}
\label{tab3}
\begin{center}
\newcommand{\tabincell}[2]{\begin{tabular}{@{}#1@{}}#2\end{tabular}}
\begin{tabular}{l|p{2.0cm}p{2.0cm}p{2.0cm}p{2.0cm}}
\hline
\bfseries Method &\bfseries 5-way-5-shot &\bfseries &\bfseries \\
\cline{2-4}
\bfseries  &\bfseries miniImageNet &\bfseries tieredImageNet &\bfseries FC100 \\
\hline \hline
EGNN \cite{kim2019edge}  & $67.33\pm0.40$ &$68.93\pm0.40$ & $47.77\pm0.42$ \\
\hline
HOSP-GNN-D & $65.75\pm0.43$ &$68.30\pm 0.40$ & $47.00\pm0.41$ \\
\hline
HOSP-GNN-S & $66.10\pm0.42$ &$68.64\pm0.41$ & $47.69\pm0.41$ \\
\hline\hline
HOSP-GNN-H & $69.19\pm0.44$ &$90.06\pm 0.30$ & $70.82\pm0.46$ \\
\hline
HOSP-GNN-H-D & $68.39\pm0.42$ &$91.11\pm0.29$ & $48.48\pm0.43$ \\
\hline
HOSP-GNN-H-S & $68.85\pm0.42$ &$91.16\pm0.29$ & $48.25\pm0.43$ \\
\hline
HOSP-GNN & $\textbf{95.98}\pm\textbf{0.21}$ &$\textbf{98.44}\pm\textbf{0.12}$ & $\textbf{70.94}\pm\textbf{0.51}$ \\
\hline
\end{tabular}
\end{center}
\end{table}

\begin{figure}
\subfigure[]{
  \centering
    \includegraphics[width=2.4in]{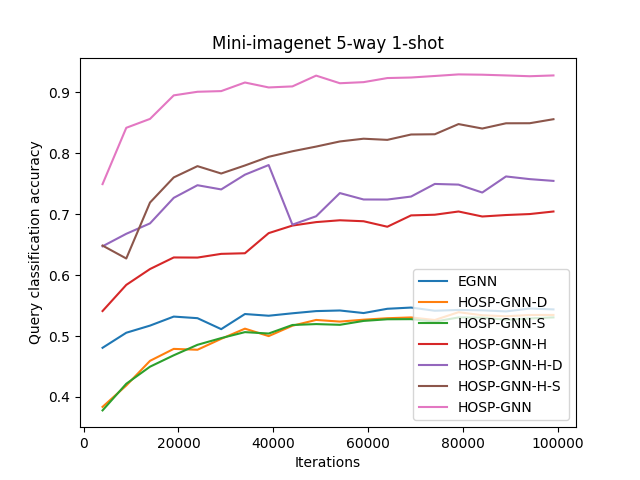}
  }
\subfigure[]{
  \centering
    \includegraphics[width=2.4in]{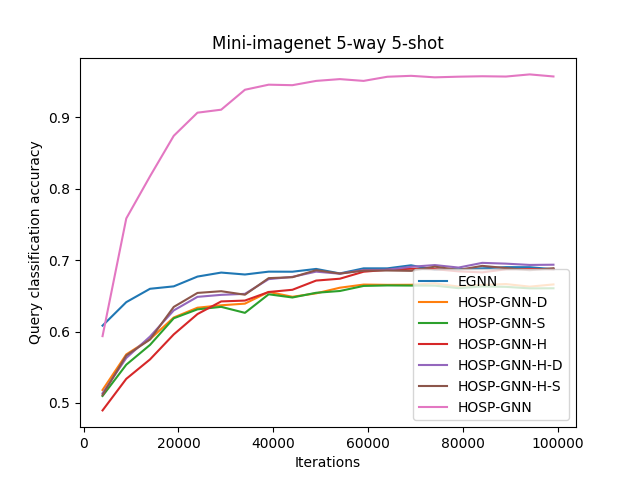}
  }
\subfigure[]{
  \centering
    \includegraphics[width=2.4in]{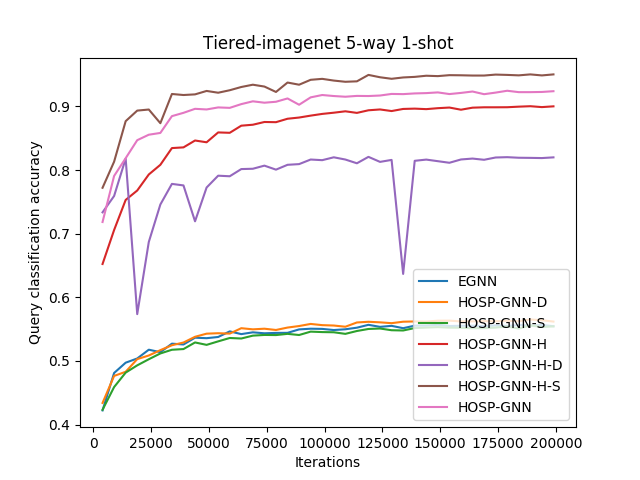}
  }
\subfigure[]{
  \centering
    \includegraphics[width=2.4in]{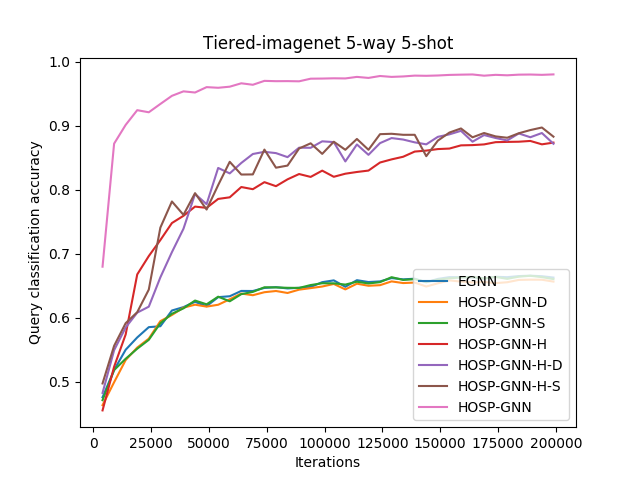}
  }
\subfigure[]{
  \centering
    \includegraphics[width=2.4in]{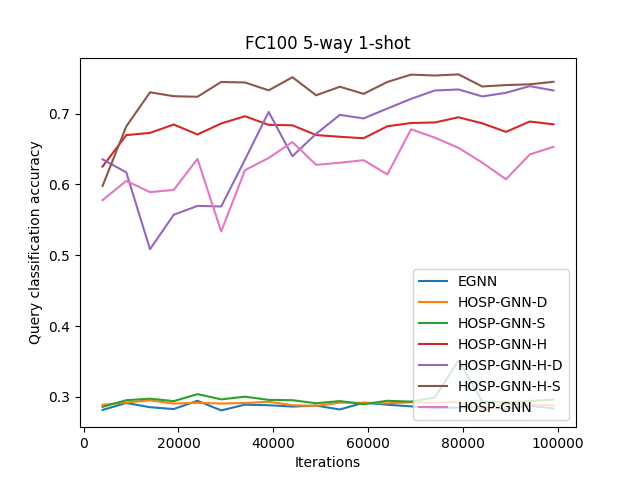}
  }
\subfigure[]{
  \centering
    \includegraphics[width=2.4in]{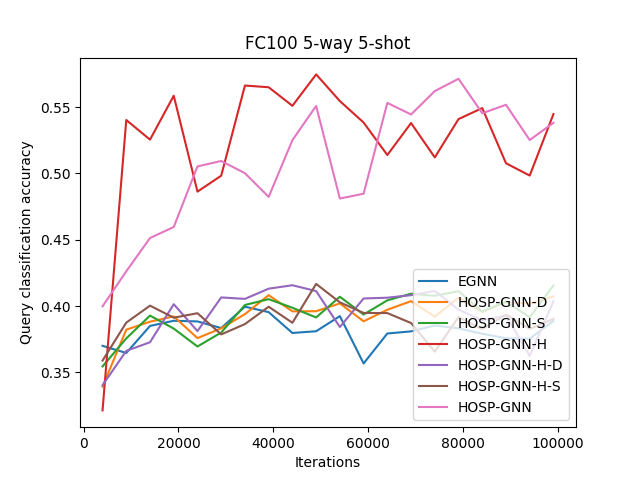}
  }
  \caption{Validation accuracy with iteration increasing for 5-way-1-shot or 5-way-5-shot in the different datasets.(a),(c) and (e) for 5-way-1-shot in miniImageNet,tieredImageNet and FC100; (b),(d) and (f) for 5-way-5-shot in miniImageNet,tieredImageNet and FC100.}
  \label{fig-5} 
\end{figure}

In Table \ref{tab2} and \ref{tab3}, the methods related the high-order structure relationship show the better performance in the base-line methods. However, the performance of HOSP-GNN based on the high-order structure combination is different because of the adaptability and coupling between the high-order structure and the pair-wise structure (similarity or dissimilarity). Figure \ref{fig-5} demonstrates the validation accuracy with iteration increasing for 5-way-1-shot or 5-way-5-shot in the different datasets. These processes also indicate the effectiveness of the high-order structure for training few-shot model. The details is analyzed in section \ref{analysis}.

\subsection{Comparison with state-of-the-arts}
In this section, we compare the proposed HOSP-GNN with the state-of-the-art methods, which include EGNN\cite{kim2019edge},MLMT \cite{fei2020meta},ArL\cite{zhang2020rethinking},
and CML-BGNN\cite{luo2019learning}, which are detailed in section \ref{ML}. These methods can capture the structure relationship of the samples in the episodic tasks based on meta-learning for few-shot learning. The difference of these method are based on the various processing ways to mine the structure relationship for few-shot models. Therefore,these methods denote the different classification performance in the benchmark datasets. Table \ref{tab4} and \ref{tab5} express that the performance of the proposed HOSP-GNN is greatly better than that of other methods. It shows that the dependence of the episodic tasks can  be better described by high-order structure based on HOSP-GNN. The detailed analysis is demonstrated in section \ref{analysis}.

\begin{table}[!ht]
\small
\renewcommand{\arraystretch}{1.0}
\caption{Comparison of HOSP-GNN method with state-of-art methods (EGNN,MLMT,ArL,and CML-BGNN) for  5-way-1-shot learning. Average accuracy (\%)of the query classes is reported in random episodic tasks.}
\label{tab4}
\begin{center}
\newcommand{\tabincell}[2]{\begin{tabular}{@{}#1@{}}#2\end{tabular}}
\begin{tabular}{l|p{2.0cm}p{2.0cm}p{2.0cm}p{2.0cm}}
\hline
\bfseries Method &\bfseries 5-way-1-shot &\bfseries &\bfseries \\
\cline{2-4}
\bfseries  &\bfseries miniImageNet &\bfseries tieredImageNet &\bfseries FC100 \\
\hline \hline
EGNN \cite{kim2019edge}  & $52.46\pm0.45$ &$57.94\pm0.42$ & $35.00\pm0.39$ \\
\hline
MLMT \cite{fei2020meta} & $72.41\pm0.49$ &$72.82\pm0.52$ & $null$ \\
\hline
ArL \cite{zhang2020rethinking} & $59.12\pm0.67$ &$null$ & $null$ \\
\hline
CML-BGNN \cite{luo2019learning} & $88.62\pm0.43$ &$88.87\pm 0.51$ & $67.67\pm1.02$ \\
\hline\hline
HOSP-GNN & $\textbf{93.93}\pm\textbf{0.37}$ &$\textbf{94.00}\pm\textbf{0.24}$ & $\textbf{76.79}\pm\textbf{0.46}$ \\
\hline
\end{tabular}
\end{center}
\end{table}

\begin{table}[!ht]
\small
\renewcommand{\arraystretch}{1.0}
\caption{Comparison of HOSP-GNN method with state-of-art methods (EGNN,MLMT,ArL,and CML-BGNN) for  5-way-5-shot learning. Average accuracy (\%)of the query classes is reported in random episodic tasks.}
\label{tab5}
\begin{center}
\newcommand{\tabincell}[2]{\begin{tabular}{@{}#1@{}}#2\end{tabular}}
\begin{tabular}{l|p{2.0cm}p{2.0cm}p{2.0cm}p{2.0cm}}
\hline
\bfseries Method &\bfseries 5-way-5-shot &\bfseries &\bfseries \\
\cline{2-4}
\bfseries  &\bfseries miniImageNet &\bfseries tieredImageNet &\bfseries FC100 \\
\hline \hline
EGNN \cite{kim2019edge}  & $67.33\pm0.40$ &$68.93\pm0.40$ & $47.77\pm0.42$ \\
\hline
MLMT \cite{fei2020meta} & $84.96\pm0.34$ &$85.97\pm0.35$ & $null$ \\
\hline
ArL \cite{zhang2020rethinking} & $73.56\pm0.45$ &$null$ & $null$ \\
\hline
CML-BGNN \cite{luo2019learning} & $92.69\pm 0.31$ &$92.77\pm 0.28$ & $63.93\pm 0.67$ \\
\hline\hline
HOSP-GNN & $\textbf{95.98}\pm\textbf{0.21}$ &$\textbf{98.44}\pm\textbf{0.12}$ & $\textbf{70.94}\pm\textbf{0.51}$ \\
\hline
\end{tabular}
\end{center}
\end{table}

\subsection{Semi-supervised few-shot learning}
In support set, we label the part of samples on all classes for the robust test of the learning model, and this situation is called semi-supervised few-shot learning. Therefore, we set $20\%$, $40\%$, and $100\%$ labeled samples of the support set for 5-way-5-shot learning in miniImageNet dataset. In this section, we compare the proposed HOSP-GNN with three graph related methods, which are GNN\cite{SatorrasE18},EGNN\cite{kim2019edge}
and CML-BGNN\cite{luo2019learning}. The common of these methods is based on graph for describing the structure of the samples, while the difference of these methods is the various ways for mining the structure of the samples. For example, GNN focuses on generic message-passing mechanism for optimizing the samples structure;EGNN emphasizes on updating mechanism for evolving the edge feature;CML-BGNN cares about the continual information of the episode tasks for structure complement; The proposed HOSP-GNN expects to mine the high-order structure for connecting the separated tasks and preserves the layer-by-layer manifold structure of the samples for constraining the model learning.The detailed analysis is indicated in section \ref{analysis}.
\begin{table}[!ht]
\small
\renewcommand{\arraystretch}{1.0}
\caption{Semi-supervised few-shot learning for  the graph related methods(GNN,EGNN,CML-BGNN and the proposed HOSP-GNN) in miniImageNet dataset. Average accuracy (\%)of the query classes is reported in random episodic tasks.}
\label{tab6}
\begin{center}
\newcommand{\tabincell}[2]{\begin{tabular}{@{}#1@{}}#2\end{tabular}}
\begin{tabular}{l|p{2.0cm}p{2.0cm}p{2.0cm}p{2.0cm}}
\hline
\bfseries Method &\bfseries miniImageNet &\bfseries 5-way-5-shot &\bfseries \\
\cline{2-4}
\bfseries  &\bfseries $20\%$-labeled &\bfseries $40\%$-labeled &\bfseries $100\%$-labeled \\
\hline \hline
GNN \cite{SatorrasE18}  & $52.45\pm0.88$ &$58.76\pm0.86$ & $66.41\pm0.63$ \\
\hline
EGNN \cite{kim2019edge} & $63.62\pm0.00$ &$64.32\pm0.00$ & $75.25\pm0.49$ \\
\hline
CML-BGNN \cite{luo2019learning} & $\textbf{88.95}\pm\textbf{0.32}$ &$\textbf{89.70}\pm\textbf{0.32}$  &$92.69\pm0.31$ \\
\hline\hline
HOSP-GNN & $65.93\pm0.38$ &$67.06\pm0.40$ & $\textbf{95.98}\pm\textbf{0.21}$ \\
\hline
\end{tabular}
\end{center}
\end{table}

\subsection{Ablation experiments for the layer number and the loss}
The proposed HOSP-GNN have two key points about structure evolution. One is the influence of the layer for model learning in graph. Another is the layer-by-layer manifold structure constraint for generating the better model with the preserved structure. Therefore, we respectively evaluate these points by ablating the part of the components from the whole model. The first experiment is about layers ablation, in which we train one layer model, two layer model and tree layer model for few-shot learning in Table \ref{tab7}. The second experiment is about the different loss, in which we set the various losses propagation for optimizing the model in Table \ref{tab8}.Table \ref{tab9} shows the parameter $\lambda$ influence to the proposed HOSP-GNN. The detailed analysis of these experimental results is shown in section \ref{analysis}.

\begin{table}[!ht]
\small
\renewcommand{\arraystretch}{1.0}
\caption{Comparison of the different layer model for  the graph related methods(GNN,EGNN,CML-BGNN and the proposed HOSP-GNN) in miniImageNet dataset. Average accuracy (\%)of the query classes is reported in random episodic tasks.}
\label{tab7}
\begin{center}
\newcommand{\tabincell}[2]{\begin{tabular}{@{}#1@{}}#2\end{tabular}}
\begin{tabular}{l|p{2.5cm}p{2.5cm}p{2.5cm}p{2.5cm}p{2.5cm}}
\hline
\bfseries Method &\bfseries miniImageNet &\bfseries 5-way-1-shot &\bfseries \\
\cline{2-4}
\bfseries  &\bfseries one layer model &\bfseries two layer model &\bfseries three layer model \\
\hline \hline
GNN \cite{SatorrasE18}  & $48.25\pm0.65$ &$49.17\pm0.35$ & $50.32\pm0.41$ \\
\hline
EGNN \cite{kim2019edge} & $55.13\pm0.44$ &$57.47\pm0.53$ & $58.65\pm0.55$ \\
\hline
CML-BGNN \cite{luo2019learning} & $\textbf{85.75}\pm\textbf{0.47}$ &$87.67\pm0.47$  &$88.62\pm0.43$ \\
\hline
HOSP-GNN & $75.13\pm0.44$ &$\textbf{87.77}\pm\textbf{0.37}$ & $\textbf{93.93}\pm\textbf{0.37}$ \\
\hline\hline
\bfseries Method &\bfseries miniImageNet &\bfseries 5-way-5-shot &\bfseries \\
\cline{2-4}
\bfseries  &\bfseries one layer model &\bfseries two layer model &\bfseries three layer model \\
\hline \hline
GNN \cite{SatorrasE18}  & $65.58\pm0.34$ &$67.21\pm0.49$ & $66.99\pm0.43$ \\
\hline
EGNN \cite{kim2019edge} & $67.76\pm0.42$ &$74.70\pm0.46$ & $75.25\pm0.49$ \\
\hline
CML-BGNN \cite{luo2019learning} & $\textbf{90.85}\pm\textbf{0.27}$ &$\textbf{91.63}\pm\textbf{0.26}$  &$92.69\pm0.31$ \\
\hline
HOSP-GNN & $67.86\pm0.41$ &$72.48\pm0.37$ & $\textbf{95.98}\pm\textbf{0.21}$ \\
\hline
\end{tabular}
\end{center}
\end{table}

\begin{table}[!ht]
\small
\renewcommand{\arraystretch}{1.0}
\caption{Comparison of the different loss model for the proposed HOSP-GNN (HOSP-GNN-loss1 for label loss in support set , and HOSP-GNN for the consideration of the label and manifold structure loss). Average accuracy (\%)of the query classes is reported in random episodic tasks.}
\label{tab8}
\begin{center}
\newcommand{\tabincell}[2]{\begin{tabular}{@{}#1@{}}#2\end{tabular}}
\begin{tabular}{l|p{2.0cm}p{2.0cm}p{2.0cm}p{2.0cm}}
\hline
\bfseries Method &\bfseries 5-way-5-shot &\bfseries  &\bfseries \\
\cline{2-4}
\bfseries  &\bfseries miniImageNet &\bfseries tieredImageNet &\bfseries FC100 \\
\hline \hline
HOSP-GNN-loss1 & $92.29\pm0.28$ &$98.41\pm0.12$ & $65.47\pm0.51$ \\
\hline\hline
HOSP-GNN & $\textbf{95.98}\pm\textbf{0.21}$ &$\textbf{98.44}\pm\textbf{0.12}$ & $\textbf{70.94}\pm\textbf{0.51}$ \\
\hline
\end{tabular}
\end{center}
\end{table}

\begin{table}[!ht]
\small
\renewcommand{\arraystretch}{1.0}
\caption{The tradeoff parameter $\lambda$ influence to few-show learning in miniImageNet.}
\label{tab9}
\begin{center}
\newcommand{\tabincell}[2]{\begin{tabular}{@{}#1@{}}#2\end{tabular}}
\begin{tabular}{l|p{1.0cm}p{1.0cm}p{1.0cm}p{1.0cm}p{1.0cm}p{1.0cm}p{1.0cm}}
\hline
\bfseries Method &\bfseries miniImageNet &\bfseries   &\bfseries 5-way-5-shot &\bfseries $\lambda$ &\bfseries \\
\cline{2-7}
\bfseries  &\bfseries $10^{-2}$ &\bfseries $10^{-3}$ &\bfseries $10^{-4}$ &\bfseries $10^{-5}$ &\bfseries $10^{-6}$ &\bfseries $10^{-7}$ \\
\hline \hline
HOSP-GNN & $94.65\pm0.22$ &$93.71\pm0.25$ & $93.43\pm0.25$ & $95.98\pm0.21$ &$95.31\pm0.21$ & $91.89\pm0.30$\\
\hline
\end{tabular}
\end{center}
\end{table}

\subsection{Experimental results analysis}
\label{analysis}
In above experiments, there are ten methods used for comparing with the proposed HOSP-GNN. In the baseline methods (HOSP-GNN,EGNN \cite{kim2019edge}, HOSP-GNN-H-S, HOSP-GNN-H-D, HOSP-GNN-H, HOSP-GNN-S and HOSP-GNN-D),we can capture the various structure information of the samples for constructing the similar learning model. In the state-of-the-art methods (EGNN\cite{kim2019edge}, MLMT \cite{fei2020meta}, ArL\cite{zhang2020rethinking},
,CML-BGNN\cite{luo2019learning} and HOSP-GNN), we demonstrate the model learning results based on the different networks framework for mining the relevance between the separated tasks. In the semi-supervised methods (GNN\cite{SatorrasE18},EGNN\cite{kim2019edge}
, CML-BGNN\cite{luo2019learning} and HOSP-GNN), we can find the labeled samples number to the performance influence for the robust testing of these methods. In ablation experiments, we build the different layers model(one layer model, two layer model and three layer model) and the various loss model (HOSP-GNN-loss1 and HOSP-GNN) for indicating their effects. The proposed HOSP-GNN can jointly consider the high-order structure and the layer-by-layer manifold structure constraints to effectively recognize the new categories. From these experiments, we have the following observations and analysis.

\begin{itemize}
\item The proposed HOSP-GNN and its Variants (HOSP-GNN-H-S, HOSP-GNN-H-D and HOSP-GNN-H) greatly outperform the base-line methods(HOSP-GNN-S, HOSP-GNN-D and EGNN) in table \ref{tab2},table \ref{tab3}, and figure \ref{fig-5}. The common characteristic of these methods (the proposed HOSP-GNN and its Variants) involves the high-order structure for learning model. Therefore,it shows that the high-order structure can better associate with the samples from the different tasks for improving the performance of few-shot learning.
\item The performance of the proposed HOSP-GNN and its Variants(HOSP-GNN-H-S, HOSP-GNN-H-D and HOSP-GNN-H) indicate the different results in the various dataset and experimental configuration in table \ref{tab2}, table \ref{tab3}, and figure \ref{fig-5}. In 5-way-1-shot learning, HOSP-GNN has the better performance than other methods in miniImageNet, while HOSP-GNN-H-S indicates the better results than others in tieredImageNet and FC100. In 5-way-5-shot learning, HOSP-GNN also shows the better performance than others in miniImageNet,tierdImageNet,and FC100. It shows that similarity, dissimilarity and high-order structure have the different influence to the model performance in the various datasets. For example, similarity,dissimilarity and high-order structure have the positive effect for recognizing the new categories in miniImageNet and tierdImageNet, while dissimilarity produces the negative effect for learning model in FC100. In any situation, high-order structure has an important and positive role for improving the model performance.
\item The proposed HOSP-GNN obviously is superior to other state-of-the-art methods in table \ref{tab4} and \ref{tab5}. These methods focus on the different aspects, which are the graph information mining based on EGNN\cite{kim2019edge}, the across task information exploitation based on MLMT \cite{fei2020meta}, the semantic-class relationship utilization based on ArL\cite{zhang2020rethinking}, the history information association based on CML-BGNN\cite{luo2019learning}, and the high-order structure exploration based on HOSP-GNN. The proposed HOSP-GNN can not only exploit the across task structure by the extension association of the high-order structure , but also use the latent manifold structure to constrain the model learning, so the proposed HOSP-GNN obtains the best performance in these methods.
\item  The proposed HOSP-GNN demonstrates the better performance than the graph related methods(GNN\cite{SatorrasE18}, EGNN\cite{kim2019edge}, and CML-BGNN\cite{luo2019learning}) based on the more labeled samples in table \ref{tab6}. The enhanced structure of the more labeled samples can efficiently propagate the discriminative information to the new categories by the high-order information evolution based on the graph. The labeled sample number has few influence on model learning based on CML-BGNN\cite{luo2019learning}. In contrast, labeled sample number has an important impact on model learning based on the graph related methods(GNN\cite{SatorrasE18}, EGNN\cite{kim2019edge}, and HOSP-GNN).
\item In the different layer model experiments, the proposed HOSP-GNN indicates the various performance with layer number changing in table \ref{tab7}. In 5-way-1-shot learning, HOSP-GNN has the better performance than other methods, while in 5-way-5-shot learning, CML-BGNN\cite{luo2019learning} shows the more challenging results than other methods. It demonstrates that layer number has an important impact on the high-order structure evolution. We can obtain the significant improvement based on the more layer model of HOSP-GNN for 5-way-5-shot in miniImageNet, while the performances of other methods almost are not changing with the layer number increasing. Therefore, the proposed HOSP-GNN trends to the more layers to exploit the high-order structure for few-shot learning.
\item In table \ref{tab8}, the different losses (supervised label loss and manifold structure loss) are considered for constructing few-shot model. HOSP-GNN (the method model based on supervised label loss and manifold structure loss) can show the better performance than HOSP-GNN-loss1(the approach involves the model with the supervised label loss). It expresses that manifold constraint with the layer-by-layer evolution can enhance the performance of model because of the intrinsic distribution consistence on the samples of the different task.

\end{itemize}

\section{Conclusion}
To associate and mine the samples relationship in the different tasks, we have presented high-order structure preserving graph neural network(HOSP-GNN) for few-shot learning. HOSP-GNN can not only describe high-order structure relationship by the relative metric in multi-samples, but also reformulate the updating rules of graph structure by the alternate computation between vertexes and edges based on high-order structure. Moreover, HOSP-GNN can enhance the model learning performance by the layer-by-layer manifold structure constraint for few-shot classification. Finally, HOSP-GNN can jointly consider similarity, dissimilarity and high-order structure to exploit the metric consistence between the separated tasks for recognizing the new categories. For evaluating the proposed HOSP-GNN, we carry out the comparison experiments about the baseline methods,the state of the art methods, the semi-supervised fashion, and the layer or loss ablation on miniImageNet, tieredImageNet and FC100. In experiments, HOSP-GNN demonstrates the prominent results for few-shot learning.
\section{Acknowledgements}
The authors would like to thank the anonymous reviewers for their insightful comments that help improve the quality of this paper. Especially, this work was supported by NSFC (Program No.61771386,Program No.61671376 and Program No.61671374), Research and Development Program of Shaanxi (Program No.2020SF-359).
\bibliography{mybibfile}

\end{document}